\RequirePackage{fix-cm}
\documentclass[twocolumn]{svjour3}          
\smartqed  

\usepackage{graphicx}
\usepackage{algorithmic}
\usepackage[ruled,vlined,commentsnumbered]{algorithm2e}
\usepackage[dvipsnames]{xcolor}
\usepackage{url}
\usepackage{array}
\usepackage{etoolbox}
\newcommand*{\affaddr}[1]{#1} 
\newcommand*{\affmark}[1][*]{\textsuperscript{#1}}
\usepackage[numbers,sort&compress]{natbib}
\usepackage{lipsum}

\begin{document} \sloppy

\title{A Deep Embedded Refined Clustering Approach for Breast Cancer Distinction based on DNA Methylation
}


\author{Roc\'{i}o del Amor\affmark[1] \and
Adri\'{a}n Colomer\affmark[1] \and
Carlos Monteagudo\affmark[2] \and
Valery Naranjo\affmark[1]}


\institute{F. Author \at
              \email{madeam2@upvnet.upv.es}           
  \\
              \affaddr{\affmark[1] Instituto de Investigaci\'{o}n e Innovaci\'{o}n en Bioingenier\'{i}a, I3B, Universitat Polit\`{e}cnica de Val\`{e}ncia, Camino de Vera s/n, 46022 Valencia, Spain.}\\
            \affaddr{\affmark[2] Pathology Department. Hospital Clínico Universitario de Valencia, Universidad de Valencia, Valencia, Spain.}\\
}

\date{Received: date / Accepted: date}

\maketitle

\begin{abstract}
Epigenetic alterations have an important role in the development of several types of cancer. Epigenetic studies generate a large amount of data, which makes it essential to develop novel models capable of dealing with large-scale data. In this work, we propose a deep embedded refined clustering method for breast cancer differentiation based on DNA methylation. In concrete, the deep learning system presented here uses the levels of CpG island methylation between 0 and 1. The proposed approach is composed of two main stages. The first stage consists in the dimensionality reduction of the methylation data based on an autoencoder. The second stage is a clustering algorithm based on the soft-assignment of the latent space provided by the autoencoder. The whole method is optimized through a weighted loss  function composed of two terms: reconstruction and  classification terms. To the best of the authors’ knowledge, no previous studies have focused on the  dimensionality reduction algorithms linked to classification trained end-to-end for DNA methylation analysis. The proposed method achieves an unsupervised clustering accuracy of 0.9927 and an error rate (\%) of 0.73 on 137 breast tissue samples. After a second test of the deep-learning-based method using a different methylation database, an accuracy of 0.9343 and an error rate (\%) of 6.57 on 45 breast tissue samples is obtained. Based on these results, the proposed algorithm outperforms other state-of-the-art methods evaluated under the same conditions for breast cancer classification based on DNA methylation data. 

\end{abstract}

\section{Introduction}
\label{introduction}

Epigenetic mechanisms are crucial for the normal development and maintenance of tissue-specific gene expression profiles in mammals. Recent advances in the field of cancer epigenetics have shown extensive reprogramming of every component of the epigenetic machinery including DNA methylation, histone modifications, nucleosome positioning and non-coding RNAs \cite{sharma2010epigenetics}. In concrete, several studies demonstrate that DNA methylation (DNAm) plays a crucial role in the tumorigenesis process \cite{zhang2015predicting,akhavan2013dna}.

\begin{figure*}[htb]
\centering
\includegraphics[width=16cm]{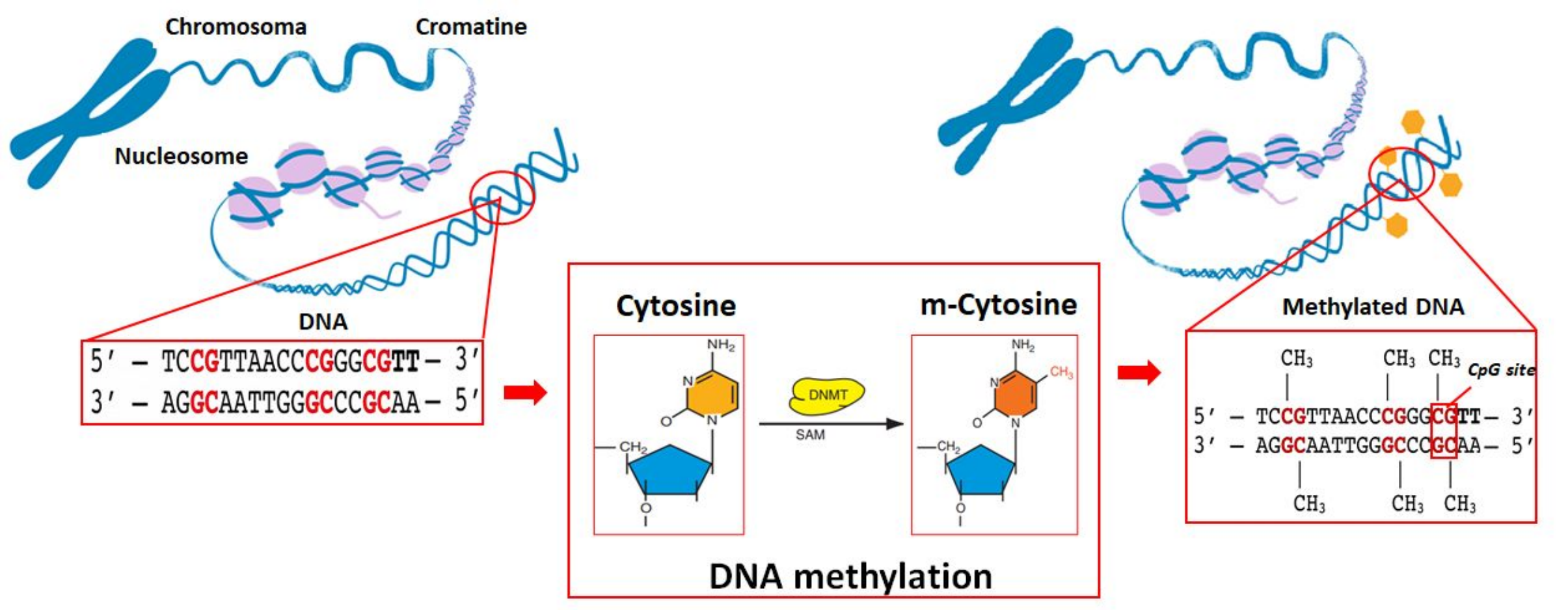}
\caption{DNA methylation process. Methylation at 5' position of the cytosine catalyzed by DNMT (DNA methyltransferases) in the presence of S-adenosyl methionine (SAM).}
\label{fig:metilation}
\end{figure*}

In mammalian cells, DNA methylation is based on the selective addition of a methyl group to the cytosine nucleotide under the action of DNA methyltransferases \cite{tsou2002dna}, Fig. \ref{fig:metilation}. Specifically, DNAm takes place in cytosines that precede guanines, known as CpG dinucleotides \cite{esteller2008epigenetics}.

\noindent CpG sites are not randomly distributed throughout the genome but there are CpG-rich areas known as CpG islands often located in the gene promoting regions. CpG islands are usually largely unmethylated in normal cells. The methylation of these CpG sites silences the promoter activity and correlates negatively with the gene expression. The methylation of the promoter regions in some vital genes, such as tumor suppressor genes, and therefore their inactivation, has been firmly established as one of the most common mechanisms for cancer development \cite{liu2019dna,du2010comparison}. Because the methylation patterns can be observed in the early stages of cancer \cite{sharma2010epigenetics}, DNA methylation analysis becomes a powerful tool in the early diagnosis, treatment and prognosis of cancer.

The DNA methylation analysis has experienced a revolution during the last decade, especially due to the adaptation of microarray technology to the study of methylation and the emergence of  Next-Generation Sequencing (NGS) \cite{martorell2019deep,bibikova2011high}. These technological advances combined with the development of techniques such as reduced representation bisulfite sequencing (RRBS), which is an efficient and high-throughput approach for analyzing the genome-wide methylation profiles,  have allowed the DNA methylation analysis at the molecular level \cite{laird2010principles}. This is the reason why, current methylation studies generate a large amount of data. Additionally, since the study of DNA methylation is still a bit expensive, the number of available samples is relatively low. 
The extremely high dimensions of the methylation data compared to the generally small number of available samples is the main limitation in the development of appropriate  methods for DNAm data analysis.
This fact makes a dimensionality reduction necessary before implementing any algorithm that identifies the presence of cancer using methylation profiles \cite{yuvaraj2013efficient}. In this context, Yuvaraj et al. presented different algorithms for dimensionality reduction based on Principal Component Analysis and Fisher Criterion \cite{yuvaraj2013efficient}. However, the DNA methylation datasets cannot be efficiently described by these dimensionality reduction methods due to its non-Gaussian character. Jazayer et al. used a non-negative matrix factorisation (NMF) for the dimensionality reduction of breast methylation data following by ELM and SVM classifiers for cancer identification. However, with the NMF algorithm, it is not possible to directly transfer the input to a smaller dimensional space than the number of samples because this method transfers the data to an output space with a dimension equal to the minimum of  \{samples, DNAm dimension\}. That is the reason why, in this study, the authors use a column-splitting method to overcome the curse of the dimensionality problem \cite{jazayeri2020breast}. 

Recent advances in the field of artificial intelligence have allowed the development of deep learning algorithms that perform an embedding of CpG methylation states to extract biologically significant lower-dimensional features \cite{si2016learning,khwaja2018deep,titus2018unsupervised}. Zhongwei et al. presented a stack of Random Boltzmann Machine (RBM) layers with the aim of reducing the dimensionality of a breast DNAm set composed of cancer and non-cancer samples. The proposed model first selected the best 5,000 features based on variance from over 27,000 features and subsequently used four RBM layers to reduce the number of features to 30. After reducing the data dimensionality, they carried out a binary classification of the generated features using unsupervised methods \cite{si2016learning}. Khwaja et al. proposed a deep autoencoder system for differentiation of several cancer types (breast cancer, lung carcinoma, lymphoblastic leukemia and Urological tumours) based on the DNA methylation states. After a statistical analysis, in which the features providing non-useful information for differentiation between cancer classes are eliminated, the authors used a Deep Belief Network for dimensionality reduction with a posterior supervised classification \cite{khwaja2018deep}. Titus et al. proposed an unsupervised deep learning framework with variational autoencoders (VAEs) to learn latent representations of the DNA methylation from three independent breast tumour datasets. They demonstrate the feasibility of VAEs to track representative differential methylation patterns among clinical sub-types of breast tumors but they do not perform any classification with the extracted characteristics \cite{titus2018unsupervised}.

Several state-of-the-art methods propose different unsupervised and supervised classification algorithms for cancer identification after performing a dimensionality reduction of the DNA methylation data \cite{liu2019dna,jazayeri2020breast,si2016learning,khwaja2018deep}. However, to the best of the author's knowledge, no previous studies have been focused on the development of both tasks simultaneously. Novel deep learning algorithms have emerged optimizing the dimensionality reduction with unsupervised classification at the same time. These methods, called deep clustering algorithms, have outperformed the state-of-the-art results for different tasks as image classification \cite{guo2017deep,xie2016unsupervised,guo2018deep}, image segmentation \cite{enguehard2019semi}, speech separation \cite{hershey2016deep, prasetio2019deep} or RNA sequencing \cite{tian2019clustering}. Therefore, our hypothesis is that since these algorithms perform well with high-dimensional data, they are likely to perform well for methylation data.

For all of the above, in this work, we proposed a deep embedded refined clustering to distinguish cancer thought DNA methylation data.  In concrete, this work is developed using two public databases containing DNAm data from breast tissues with and without cancer. The proposed method is composed of an autoencoder to carry out the dimensionality reduction followed by a soft-assignment algorithm to perform an unsupervised classification. This algorithm is end-to-end trained to accomplish the data classification while optimising the dimensionality reduction. As the main novelty, the method is optimized through a weighted loss function. This loss function is composed of two terms: (1) a reconstruction term in charge of  optimizing the latent features provided by the autoencoder algorithm and (2) a clustering term  used to improve the classification based on the latent features of the autoencoder. To the best of the authors’ knowledge, no previous studies have addressed the distinction of cancer based on DNAm using an end-to-end trained dimensionality reduction and classification method. The proposed method is widely validated and compared to the use of autoencoder and variational autoencoder for dimensionality reduction with a subsequent unsupervised classification. 

The rest of the paper is organised as follows: in Section \ref{ssec:material} we introduce the databases used in this work, DNAm sets containing the methylation level (between 0 and 1) of different CpG region related to cancer. In Section 3, we describe the methodology. In particular, Section \ref{3.1}  describes the statistical  analysis performed on the DNAm data, Section \ref{autoencoders} presents the dimensionality reduction algorithms used in this work, conventional  and variational autoencoder, and Section \ref{DERC}  describes the details of the proposed deep clustering method. In Section \ref{4}, we describe the performed experiments in order to validate our method and in Section \ref{5} we discuss the results obtained. Finally, Section  \ref{6} summarises the conclusions extracted with the carried out experiments.

\section{Materials}
\label{ssec:material}

For this study, we used two methylation datasets obtained from Gene Expression Omnibus (GEO) website \cite{GEO}. GEO is a public functional genomics data repository supporting MIAME-compliant data submissions. Specifically, we used the GSE32393 and the GSE57285 series to evaluate the proposed method. Note that the methodology proposed in this work was applied on the first series. The second series was used as an external database to perform an additional test and demonstrate the robustness of the proposed methodology. The GSE50220 series is composed of breast tissue samples from 114 breast cancers and 23 non-neoplastic breast tissues. The breast cancer tissue samples come from women (mean age 59.4) who were diagnosed with breast cancer. Among the cancers, 33 were at stage 1 and 81 at stage 2/3/4. All 23 non-neoplastic samples are from healthy women (mean age 47.6). The GSE50220 series is composed of breast tissue samples from 36 breast cancer and 9 normal control. Among the breast cancer, 20 were non-irradiated breast cancer and the rest were irradiated tumors.  In all cases, to obtain the methylation data, the Illumina Infinium 27k Human DNA methylation Beadchip v1.2 was used at approximately 27,000 CpGs from women with and without breast cancer. For each sample, 27,578 DNA methylation profiles were obtained. The methylation status of each CpG site varies from 0 to 1. Under ideal conditions, a value of 0 means the CpG site is completely unmethylated and the value of 1 indicates the site is fully methylated.

\section{Methodology}
\subsection{Statistical analysis}
\label{3.1}

To conduct the prescreening procedure and obtain the methylation sites with the most differential methylation expression, a previous statistical analysis of the CpG methylation data was carried out. First, a hypothesis contrast to analyze the level of independence  between  pairs of variables was performed. For this purpose, the correlation coefficient \textit{$\rho$} and the \textit{p-value} of the correlation matrix were calculated to remove those variables that meet both \textit{p-value} $\leq$ $\alpha$ and $|\rho|$ $\leq$ $0.90$, being $\alpha$ the level of significance with a value of 0.05 for this application. After that, we performed different contrasting hypotheses to analyze the discriminatory ability of each variable regarding the class. Depending on if the variables fit a normal distribution or not, the hypothesis test performed was the t-student or the Wilcoxon Rank-Sum, respectively.  After the statistical analysis, we reduced the 27,578 DNA methylation features of the  GSE32393 series to 10,153. These features were the input for the following stage.

\subsection{Dimensionality reduction}
\label{autoencoders}

In order to explore the well-known non-supervised algorithms to reduce the data dimensionality based on deep learning techniques, the conventional and the variational autoencoder were tested. In this section, we detail the characteristics of both algorithms as well as their main differences.

\begin{itemize}
    \item \textit{Conventional autoencoder}
\end{itemize}

\begin{figure}[htb]
\centering
\includegraphics[width=8cm]{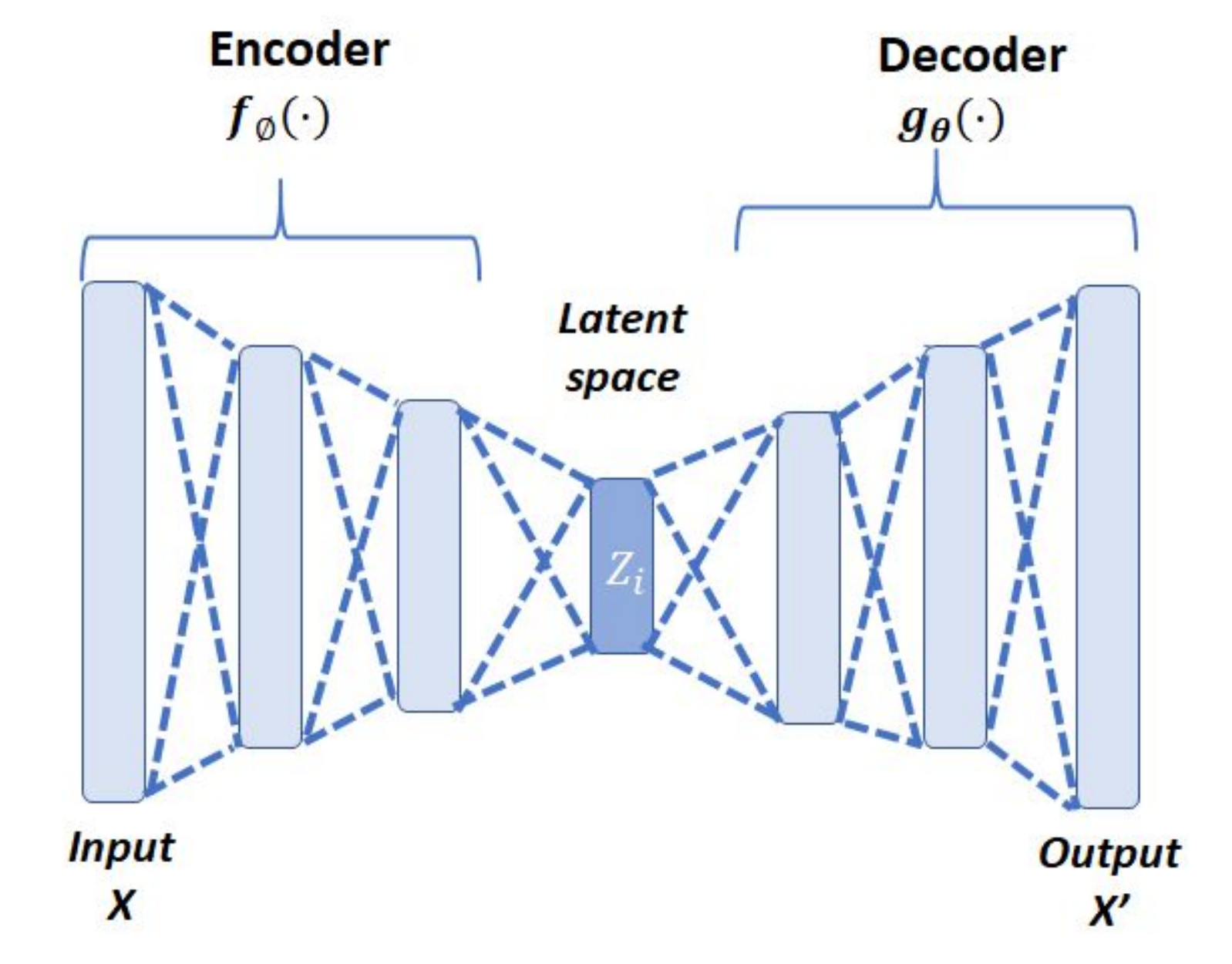}
\caption{Architecture of the proposed conventional autoencoder used for the non-supervised dimensionality reduction. }
\label{fig:autoencoder}
\end{figure}

Autoencoder (AE) is one of the most significant algorithms in unsupervised data representation. The objective of this method is to train a mapping function to ensure the minimum reconstruction error between input and output \cite{min2018survey}. As it can be observed in Fig. \ref{fig:autoencoder}, the conventional autoencoder architecture is composed mainly of two stages: the encoder and the decoder stages. The encoder step is in charge of transforming the input data $\textbf{X}$ into a latent representation $\textbf{Z}$ through a non-linear mapping function, $\textbf{Z}=f_\phi(\textbf{X})$, where $\phi$ are the learnable parameters of the encoder architecture. The dimensionality of the latent space $\textbf{Z}$ is much smaller than the corresponding input data to avoid the curse of dimensionality \cite{xie2016unsupervised}. Since the latent space is a non-linear combination of the input data  with smaller dimensionality, it can represent the most salient features of the data. The decoder stage produces the reconstruction of the data based on the features embedded in the latent space, $\textbf{R}=g_\theta(\textbf{Z})$. The reconstructed representation \textbf{R} is required to be as similar to $\textbf{X}$ as possible. Therefore, given a set of data samples $\textbf{X}=\left\lbrace x_i,...,x_n \right\rbrace$, being $n$ the number of available samples, the autoencoder model is optimized with the following formula:

\begin{equation}
\label{reconstruction_error}
    	\min_{\theta,\phi}L_{rec}=\min \frac{1}{n}\sum_{i=1}^{n} ||x_{i}-g_\theta(f_\phi(x_i))||^2
\end{equation}

\noindent where $\theta$ and $\phi$ denote the parameters of encoder and decoder, respectively.

The autoencoder architecture can vary between a simple multilayer perceptron (MLP), a long short-term memory (LSTM) network or a convolutional neural network (CNN), depending on the use case. In case the input data is 1-D and unrelated in time, both the encoder and decoder  are usually constructed by a multilayer perceptron.

\begin{itemize}
    \item \textit{Variational autoencoder}
\end{itemize}

\begin{figure*}[hbt]
\centering
\includegraphics[width=10cm]{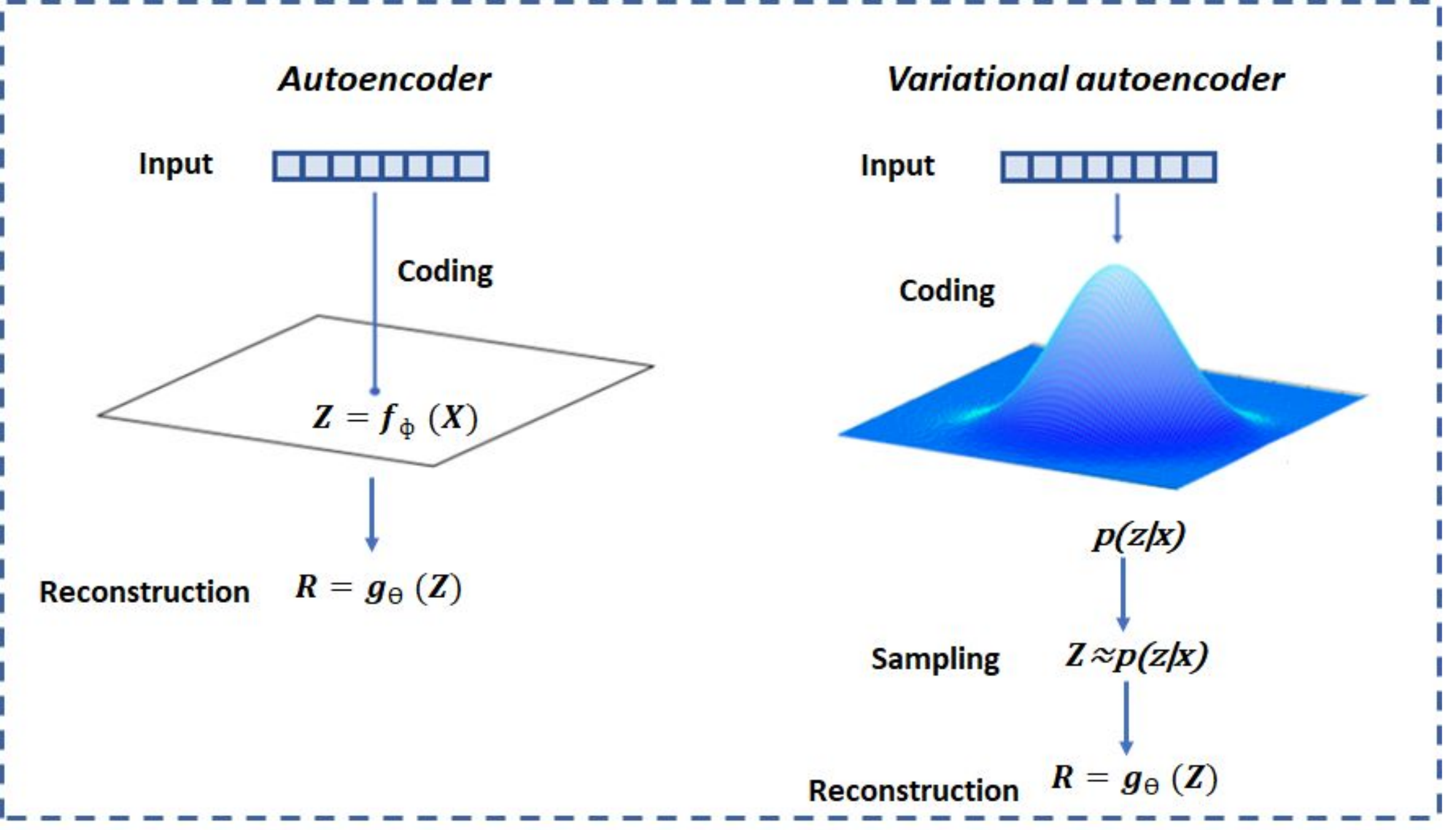}
\caption{Main differences between a conventional and a variational autoencoder. Instead of just learning a function representing the data (a compressed representation) like conventional autoencoders, variational autoencoders learn the parameters of a probability distribution representing the input data.}
\label{fig:difference}
\end{figure*}

Variational autoencoder (VAE) is an unsupervised  approach composed also of an encoder-decoder architecture like the conventional autoencoder aforementioned \cite{titus2018unsupervised}. However, the main difference between a conventional and a variational autoencoder lies in the fact that the VAE introduces a regularisation into the latent space to improve its properties. With a VAE, the input data is coded as a  normal multivariate distribution $p(z|x)$ around a point in the latent space. In this way, the encoder part is optimized to obtain the mean and covariance matrix of a normal multivariate distribution, see Fig. \ref{fig:difference}.

\begin{figure*}[hbt]
\centering
\includegraphics[width=12cm]{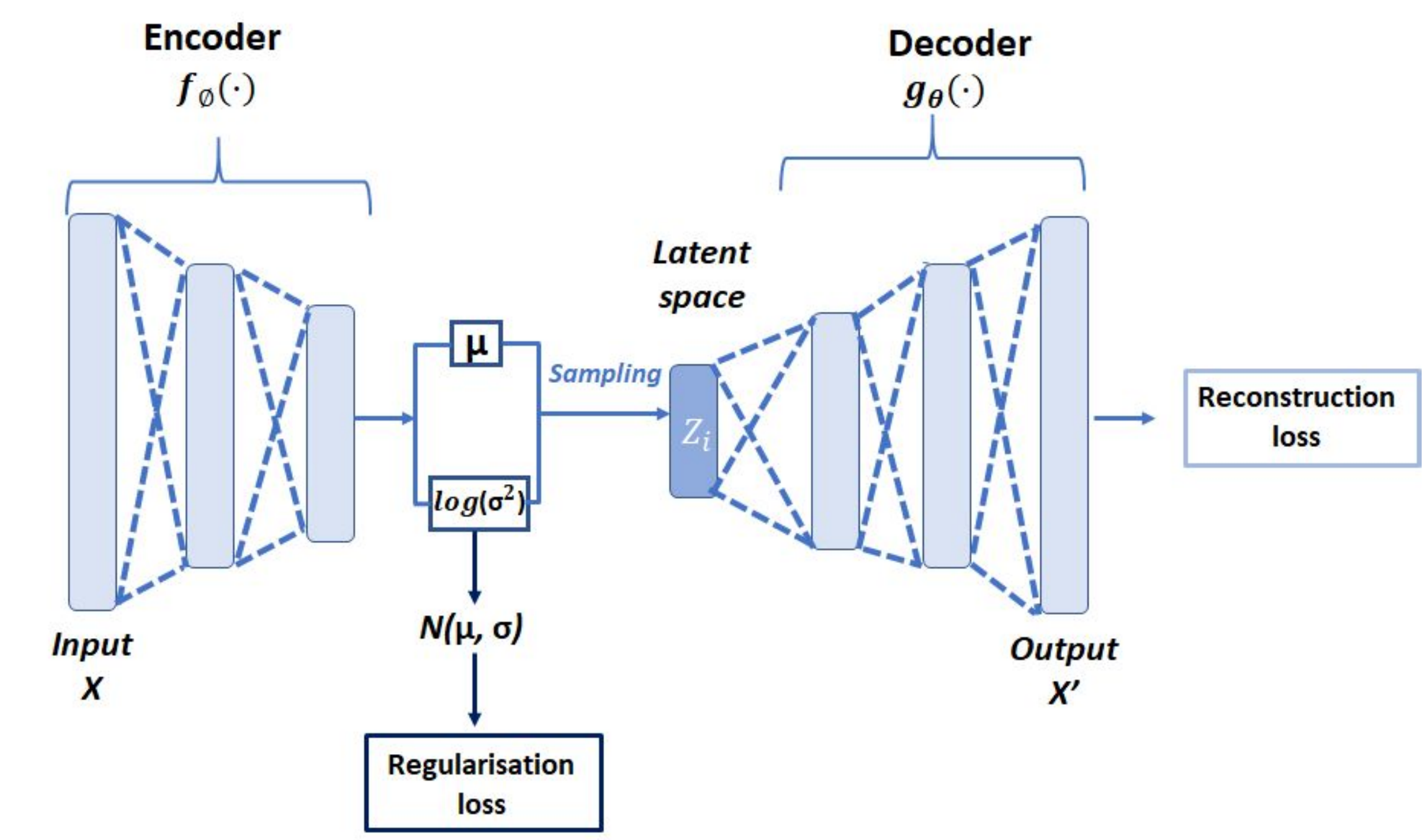}
\caption{Architecture of a variational autoencoder. The proposed algorithm is optimized by minimizing two loss functions. One of them corresponding to the latent space regularisation and the other one corresponding to the input data reconstruction.}
\label{fig:variational_autoencoder}
\end{figure*}

The VAE algorithm assumes that there is no correlation between any latent space dimensions and, therefore, the covariance matrix is diagonal. In this way, the encoder only needs to assign each input sample to a mean and a variance vectors. In addition, the logarithm of the variance is assigned, as this can take any real number in the range $(-\infty, \infty)$, matching the natural output range from a neural network, whereas that variance values are always positive, see Fig. \ref{fig:variational_autoencoder}. 

In order to provide continuity and completeness to the latent space, it is necessary to regularize both the logarithm of the variance and the mean of the distributions returned by the encoder. This regularisation is achieved by matching the encoder output distribution to the standard normal distribution ($\mu=0$ and $\sigma=1$).

After obtaining and optimizing the parameters of mean and variance of the latent distributions, it is necessary to take samples of the learned representations to reconstruct the original input data. Samples of the encoder output distribution are obtained as follows:

\begin{equation}
\centering
    Z \approx p(z|x)=\mu + \sigma \cdot \epsilon
\end{equation}

\noindent where $\epsilon$ is randomly sampled from a standard normal distribution and $\sigma=\exp(\frac{\log(\sigma^2)}{2})$.

The minimized loss function in a variational autoencoder  is composed of two terms: (1) a reconstruction term that compares the reconstructed data to the original input in order to get as effective encoding-decoding as possible and (2) a regularisation term in charge of regularizing the latent space organization, Fig. \ref{fig:variational_autoencoder}. The regularisation term is expressed as the \textit{Kulback-Leibler} (KL) divergence that measures the difference between the predicted latent probability distribution of the data and the standard normal distribution in terms of mean and variance of the two distributions \cite{variational_autoencoder}:
\begin{equation}
D_{KL}[N(\mu,\sigma)||N(0,1)]=\frac{1}{2}\sum(1+log(\sigma^2)-\mu^2-\sigma^2)
\end{equation}
The \textit{Kulback-Leibler} function is minimised to 0 if $\mu = 0$ and $log(\sigma^2)=0$ for all dimensions. As these two terms begin to differ from 0, the variational autoencoder loss increases. The compensation between the reconstruction error and the KL divergence is a hyper-parameter to be adjusted in this type of architecture.

\subsection{Proposed method: Deep embedded refined clustering}
\label{DERC}

\begin{figure*}[htb]
\centering
\includegraphics[width=13cm]{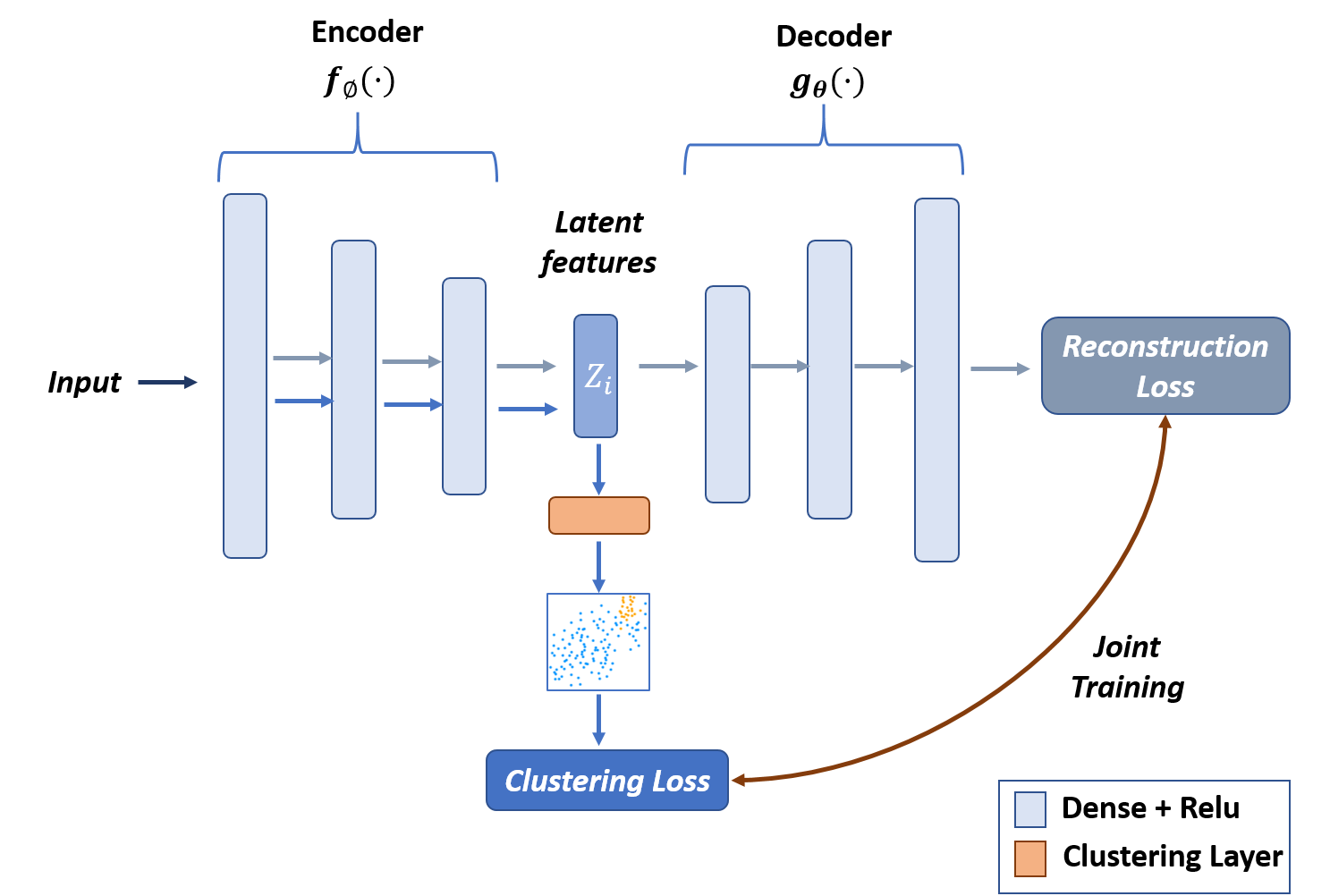}
\caption{Architecture of the proposed method (DERC) to detect breast cancer using DNA methylation data. The proposed algorithm is trained minimizing both, clustering and reconstruction loss.}
\label{fig:deepC}
\end{figure*}

Once the data dimensionality is reduced, we classify the samples in cancerous and non-cancerous. Reducing the data dimensionality without information about the different subjacent data distributions weakens the representativeness of the embedded features concerning the class and thereby, the performance of the subsequent classification worsens. For this reason, we consider that dimensionality reduction and classification should be optimized at the same time. In this context, we propose a deep refined embedded clustering (DERC) approach for classifying the DNA methylation data, see Fig. \ref{fig:deepC}. It is composed of an autoencoder in charge of the dimensionality reduction and a cluster assignment corresponding to the unsupervised classification stage (clustering layer in Fig. \ref{fig:deepC}). This approach is trained end-to-end optimizing the dimensionality reduction and classification in the same step and not in two different steps as all the algorithms proposed  for DNA methylation analysis in the literature. 

During the training process, the encoder and decoder weights of the autoencoder,  $W$ and $W'$ respectively, are updated in each iteration in order to refine the latent features of the encoder output \textbf{Z}. The proposed clustering layer (linked to the encoder output) obtains the soft-assignment  probabilities $q_{i,j}$ between the embedded points $z_i$ and the cluster centroids $\left\lbrace \mu_j\right\rbrace_{j=1}^k$ every $T$ iterations, being $k$ the number of cluster centroids. The soft-assignment probabilities ($q_{i,j}$) are obtained with the Student's t-distribution proposed in \cite{xie2016unsupervised}. Using $q_{i,j}$, the target probabilities $p_{i,j}$ are updated, see Algorithm \ref{DECR_algorithm}. These target probabilities allow the refinement of the cluster centroids by learning from the current high-confidence assignments. To take into account the refinement of the latent space carried out by autoencoder while the samples are classified in one of the two clusters (cancer and non-cancer), the proposed model is trained end-to-end minimizing both reconstruction ${L_{rec}}$ and clustering loss ${L_{cluster}}$ terms: 

\begin{equation}
    L=L_{cluster}+ \beta L_{rec}
\end{equation}

\noindent where $\beta$ balances the importance of the losses due to the reconstruction of the data. The term $L_{rec}$  was defined in Equation (\ref{reconstruction_error}) and it is minimized to obtain the maximum similarity between the input and the output data improving the representation of the latent space. $L_{cluster}$ is defined by the Kullback–Leibler (KL) divergence loss between the soft-assignments and the target probabilities, $q_{i,j}$ and $p_{i,j}$ respectively:

\begin{equation}
    L_{cluster}=\sum_{i}\sum_{j} p_{i,j} log \frac{p_{i,j}}{q_{i,j}}
\end{equation}

\noindent The clustering term is minimized to achieve the soft-assignments $q_{i,j}$ and the target $p_{i,j}$ probabilities to be as similar as possible. In this way, the centroids are refined and the latent space obtained by the autoencoder is regularized to achieve a correct distinction between breast cancer and non-breast cancer samples. As discussed above, the  hyper-parameter $\beta$ balances the importance of losses due to the data reconstruction. If $\beta$ is high, the  data reconstruction term will predominate and the classification between cancerous and non-cancerous samples will worsen. Otherwise, if this term is too low, the reconstruction losses will be marginal and the features of the latent space will not be optimized correctly. Consequently, the latent features will be  very different from the input data, decreasing the accuracy of the classification. Therefore, $\beta$ is a hyper-parameter that needs to be properly adjusted. In  Step 2 of Algorithm \ref{DECR_algorithm}, the methodology used to optimize the proposed DERC algorithm is detailed.

Note that to train the proposed method, a previous initialization of the centroids with latent characteristics is necessary (Step 1 of the Algorithm \ref{DECR_algorithm}). In the experimental section, we present an experiment (Section \ref{autoencoder_variationalAuto}) aimed at determining which of the dimensionality reduction models is optimal for this initialization.

\begin{algorithm}[ht]
\caption{Proposed methodology for the DERC approach.}
\label{DECR_algorithm}
\footnotesize
\small
\BlankLine
\SetKwInput{KwInput}{Input}                
\SetKwInput{KwOutput}{Output} 

\KwInput {Methylation data \textbf{X}; number of clusters $k$; update interval $T$; batch-size $bs$; learning rate $lr$;  number of samples $n$.} 
\KwOutput {Cluster assignment $\left\lbrace c_i\right\rbrace_{i=1}^n$ of each methylation sample $\left\lbrace x_i\right\rbrace_{i=1}^n$. \BlankLine}

\textbf{Step 1: Previous data dimensionality reduction} 

(1) Pre-train the proposed autoencoder algorithm.

(2) Obtain the centroid initialisation $\left\lbrace \mu_j\right\rbrace_{j=1}^k$ by running K-means on the latent space \textbf{Z} of the pre-trained autoencoder.

\textbf{Step 2: Clustering using the proposed DERC method}

\textit{End-to-end DERC optimization: \\} 
\For{$ ite \leftarrow 1$ $\KwTo$ $\mathbf{MAXiter}$}{ 
    \textit{\%Choose a batch of samples  \textbf{$X_{bs}$} $\subset$ \textbf{X} \\} 
  \If{$ite\% T==0$}
    {
    $z_i \leftarrow f_\phi({x_i})$, $ \forall x_i \in$  \textbf{X}\;
    update $q_{i,j} \leftarrow \frac{(1+||z_i+\mu_j||^2)^{-1}}{\sum_{j'}(1+||z_i+\mu_{j´}||^2)}$, $j\not =j'$\;
    update $p_{i,j} \leftarrow \frac{q_{i,j}^2/f_j}{\sum_{j'}q_{i,j'}^2/f_{j'}}$, $f_{j}=\sum_{i}q_{i,j}$\;} 
    \BlankLine
    \textit{\% Update encoder weights $W$, decoder weights $W'$ and centroids $\left\lbrace \mu_j\right\rbrace_{j=1}^k$:}\\
    
    update $ W \leftarrow W- \frac{lr}{bs} \sum_{i=1}^{bs} \left[ \beta \frac{\partial L_{rec}}{W'} + \frac{\partial L_{cluster}}{\partial W} \right]; $
    
    update $W'\leftarrow W'- \frac{lr}{bs} \sum_{i=1}^{bs}\frac{\partial L_{rec}}{W'};$
    
    update $\mu_j \leftarrow\mu_j-\frac{lr}{bs}\sum_{i=1}^{bs}{\frac{\partial L_{cluster}}{\partial  \mu_j}}$;
    }
    \textit{Final prediction stage: \\} 
    \For{$i \leftarrow 1$ $\KwTo$ $\mathbf{n}$}{
    $c_i=argmax_j(q_{i,j})$  \\}

\end{algorithm}

\section{Experimental results}
\label{4}

As we mentioned in Section \ref{ssec:material},
the DNA methylation databases used were obtained from the Gene Expression Omnibus (GEO) website. In this section, we used the dataset GSE32393 to evaluate the dimensionality reduction and the unsupervised deep clustering performance. The dataset GSE50220 was used as an external validation to demonstrate that the proposed method can generalise to other breast methylation databases. It should be noticed that all experiments were performed on an Intel i7 @ 3.10 GHz of 16 GB of RAM with a Titan V GPU of 12 GB of RAM. The proposed methods were executed in Python 3.5 using TensorFlow 2.0. 

\subsection{GSE32393 Series:  Performance evaluation}

\subsubsection{Dimensionality reduction and classification separately}
\label{autoencoder_variationalAuto}
As mentioned above, an initial latent space with a lower dimensionality than the input data for the cluster centroid initialisation is necessary. In this section, we detail  a comparison between the latent space obtained using the conventional and the variational autoencoder and the classification results after applying the K-means algorithm on each latent space. In this way, it will be demonstrated which  algorithm is the most suitable for dimensionality reduction in the end-to-end proposed method. 

\begin{figure}[hbt]
\centering
\includegraphics[width=8.7cm]{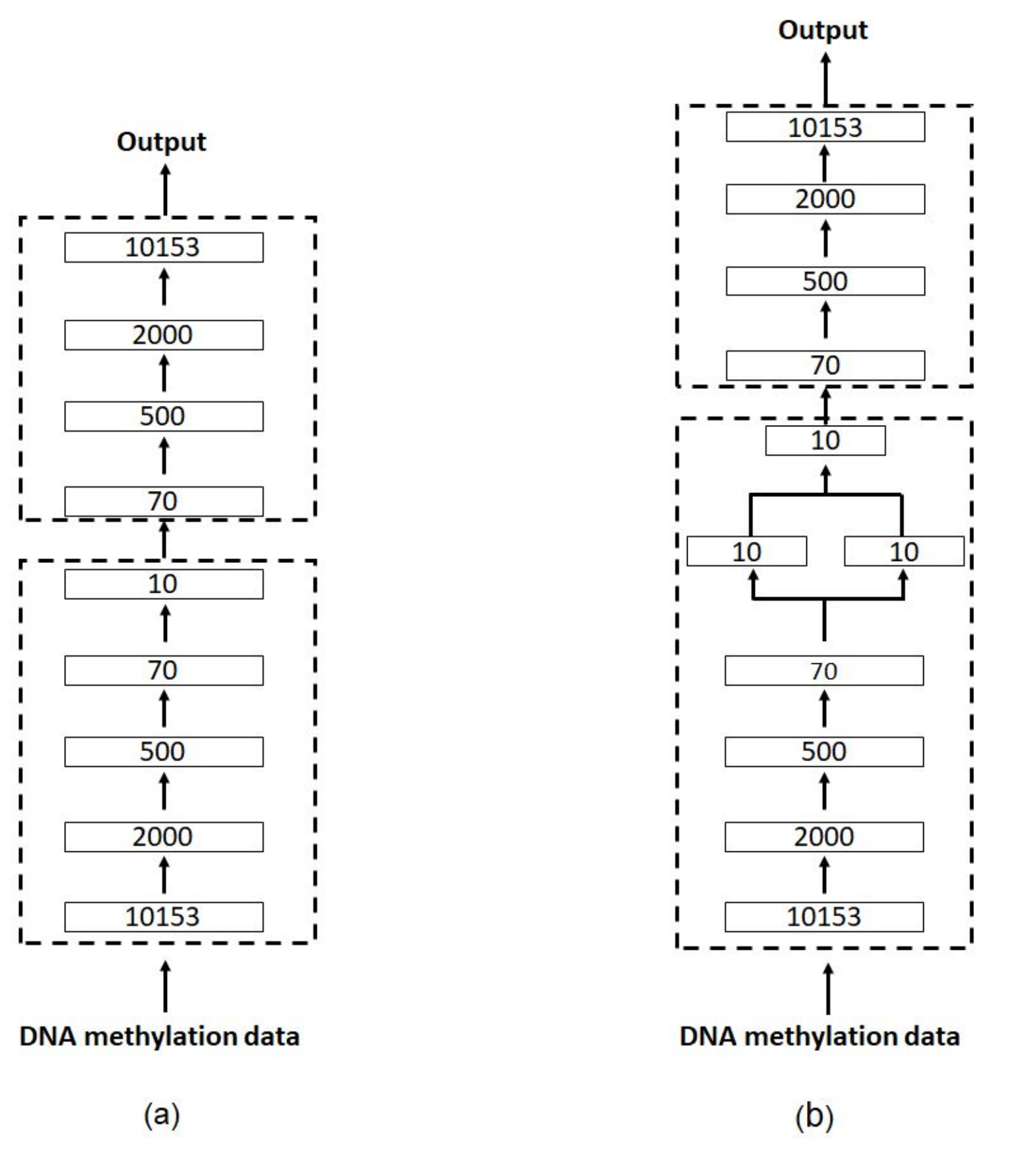}
\caption{Final dimensionality reduction architectures. (a) Conventional autoencoder architecture composed of 4 stacks. (b) Variational autoencoder architecture composed of 4 stacks. Note that with the variational autoencoder algorithm, the latent space is obtained in two stages, two dense layers of 10 neurons representing the mean and the logarithm of the variance of the latent distribution and a sampling layer to obtain the points of the latent space.}
\label{fig:compration arcchitecture}
\end{figure}

\textbf{Ablation experiment}. The 10,153  CpG sites obtained after statistical analysis of the raw methylation data were the input of the proposed dimensionality reduction algorithms, conventional and variational autoencoders. In both cases, the dimensionality reduction was carried out using an architecture composed of 4 stacks. The number of neurons (input, output) of the 3 top layers were set to \{(10153, 2000), (2000, 500), (500, 70)\}, respectively (see Fig. \ref{fig:compration arcchitecture}). These layers were composed of a dense layer with ReLU as activation function except for the last decoder layer that was constituted of the sigmoid function in order to obtain an output value between 0 and 1, range of the methylation data values. The kernel weights were initialised with random numbers drawn from a uniform distribution within $[-l,l]$, where $l=\sqrt{3 \cdot s/n_{input}}$, being $s=\frac{1}{3}$ and $n_{input}$ the number of input units. The top layer output (latent space dimension) was set to \{10, 20, 30\}. Note that, these settings were obtained from empirical evaluations with a wide range of settings and we use only the best parameters here. After intense experiments, the optimal dimension of the latent space for both algorithms turned out to be 10 neurons.

To show the performance of AE and VAE and to demonstrate that they are not over-adjusted to the data, a first experiment was performed using 10\% of the GSE32393 database as a validation set and the rest 90\% as a training set. Subsequently, both algorithms were trained using the whole database (Entire prediction). The optimal hyper-parameters combination was achieved by training both algorithms during 300 epochs, using the Stochastic Gradient Descent (SGD) optimizer with a learning rate of 1 and a batch-size of 8. Regarding the loss function, in the case of the conventional autoencoder, the mean square error (MSE) was used. However, the variational autoencoder loss function was composed of two terms: MSE weighted by 0.8 and Kullback–Leibler
(KL) divergence.

After training the dimensional reduction algorithms with the entire GSE32393 series and obtaining the features in the embedding space (encoder output), the classification results were obtained using K-means. To achieve this classification, we ran K-means with 80 restarts and selected the best solution.

\textbf{Qualitative and quantitative results}. 
After training the proposed dimensionality reduction algorithms, the results in terms of reconstruction error for both autoencoders are shown in Table \ref{tab:reconstruction_error}. 

\begin{table}[hbt]
\centering
\small
\caption{Reconstruction error of the proposed dimensionality reduction algorithms. Conventional autoencoder (AE) and variational autoencoder (VAE).}
\label{tab:reconstruction_error}
\resizebox{8cm}{!}{%
\begin{tabular}{llll} 

\hline
\textbf{Method} & \multicolumn{3}{c}{\textbf{Reconstruction Loss}} \\ \hline
                & Training    & Validation    & Entire prediction   \\ \hline
\textbf{AE}              & \textbf{0.0062}           & \textbf{0.0054}            & \textbf{0.0057}                  \\ \hline
VAE             & 0.0082          & 0.0074             & 0.0082                  \\ \hline
\end{tabular}%
}
\end{table}

\begin{figure*}[ht]
\begin{center}
\begin{tabular}{cc}

\includegraphics[width=6.6cm, height=4.8cm]{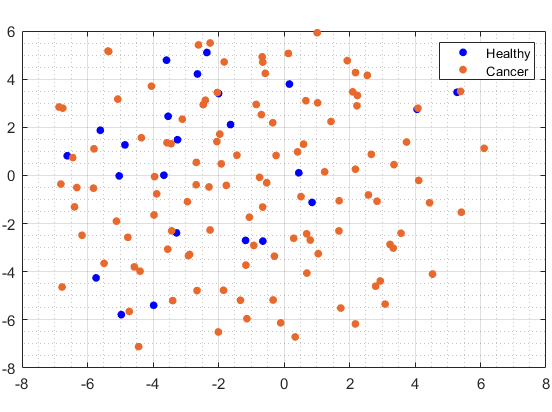} &
\includegraphics[width=6.6cm, height=4.8cm]{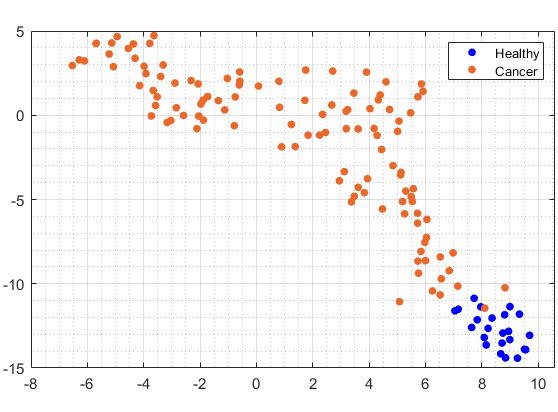}

\\
(a) &
(b)  

\\
\end{tabular}
\end{center}
\caption{ Latent space of the dimensionality reduction algorithms. (a) Visualisation of 10-dimensional features extracted by the latent space of the pre-trained variational autoencoder. (b) Visualisation of 10-dimensional features extracted by the latent space of the pre-trained conventional autoencoder. }
\label{data_visualization}
\end{figure*}

In order to visualise in a qualitative way the effect of the tested dimensionality reduction methods (AE and VAE) over the data distribution, we used the t-distributed stochastic neighbour embedding (t-SNE) method to represent the latent space into a two-dimensional space. T-SNE is a nonlinear dimensionality reduction technique that embeds high-dimensional data into a space of two or three dimensions, which can then be visualised by a scatter plot \cite{maaten2008visualizing}. In Fig. \ref{data_visualization}, we show the representation of the data (latent space of the pre-trained autoencoder (a) and latent space of the pre-trained variational autoencoder (b)) in a two-dimensional space. 

To quantitatively evaluate the performance of the clustering assignments, several metrics were computed: the unsupervised clustering accuracy (ACC), the error rate (\%) and the false positive (FP) and the  false negative (FN) ratios. The ACC metric \cite{min2018survey} is defined as follows:

\begin{equation}
    ACC=max_m  \left(\frac{\sum_{i=1}^{n}1 {\left\lbrace y_i=m(c_i)\right\rbrace}}{n}\right)
\end{equation}

\noindent where $y_i$ is the ground-truth label, $c_i$ is the cluster assignment generated by the algorithm, and $m$ is a mapping function which ranges over all possible one-to-one mappings between assignments and labels. The error rate ($\%$) is calculated according to the following formula:

\begin{equation}
Error \; rate \;(\%)= (1-ACC) \cdot 100
\end{equation}

In Table \ref{tab:results_met}, the above-mentioned metrics were calculated for the input data + K-means clustering, the latent space of the pre-trained autoencoder(AE)+K-means and the latent space of the pre-trained variational autoencoder (VAE)+ K-means. Note that the input data is referred to the 10,153 features extracted after the statistical analysis.

\begin{table*}[ht]
\centering
\caption{Comparison of the K-means clustering effect based on different feature extraction.}
\label{tab:results_met}
\resizebox{10 cm}{!}{%
\begin{tabular}{lcccc}
\textbf{Method} & \textit{\textbf{ACC}} & \textit{\textbf{Error Rate(\%)}} & \textit{\textbf{FP}} & \textit{\textbf{FN}} \\ \hline
\multicolumn{1}{c}{Input Data + K-means} & 0.6715 & 32.85 & 45 & 0 \\ \hline
AE + K-means                             & \textbf{0.9343} & \textbf{6.57}  & \textbf{9}  & \textbf{0} \\ \hline
VAE + K-means                         & 0.5693 & 43.07  & 50  & 28 \\ \hline
\end{tabular}%
}
\end{table*}

\subsubsection{ Dimensionality reduction and classification jointly (DERC)}

After initializing the centroids using the algorithm with the lowest losses and the better prediction when K-means was used, in this case the conventional autoencoder, the deep embedded refined clustering algorithm was trained. As we explained in Section \ref{DERC}, the $\beta$ value, which weights the terms that composed the loss function of the DERC algorithm, is an important parameter to adjust. For this reason,  we develop in this section a comparison between different $\beta$ values exposing their influence on the clustering assignment.

\textbf{Ablation experiment}. After pre-training the conventional autoencoder model (with the parameters detailed in Section \ref{autoencoder_variationalAuto}), we added the clustering layer to the output of the autoencoder latent space, see Table \ref{CNN_from_scratch} for the layer layout in the final architecture. 

\begin{table}[htbp]
\caption{Architecture of the proposed deep embedded refined clustering model.}
\label{CNN_from_scratch}
\renewcommand{\arraystretch}{0.7} 
\setlength\tabcolsep{4.5 pt} 
\small
\begin{center}
\begin{tabular}{ccc}
\hline
 \textbf{Layer name} & \textbf{Output shape}                & \textbf{Connected to}            \\ \hline
Input\_layer         & 10153                                 & N/A                             \\
Encoder\_0          & 2000                                  & Input\_layer                      \\
Encoder\_1          & 500                                   & Encoder\_0                      \\ 
Encoder\_2          & 70                                    & Encoder\_1                       \\ 
Encoder\_3          & 10                                    & Encoder\_2                      \\ 
Decoder\_3          & 70                                    & Encoder\_3                      \\
Decoder\_2          & 500                                   & Decoder\_3                       \\ 
Decoder\_1          & 2000                                  & Decoder\_2                     \\
Clustering\_layer     & 2                                     & Encoder\_3                             \\ 
Decoder\_0          & 10153                                 & Decoder\_1                             \\ \hline
\end{tabular}
\end{center}
\end{table}

In order to evaluate the $\beta$ value on the performance of the clustering algorithm, we kept the rest of  the hyper-parameters constant during the different experiments. In particular, the entire deep embedding clustering method was optimised by stochastic gradient descent (SGD) with a learning rate of 0.01 and a momentum of 0.9. The proposed method was trained during 50 epochs using a batch-size of 8 samples and the target distribution of the clustering layer was updated every 10 iterations. Note that these hyper-parameters were obtained from empirical  evaluations  with  a  wide range  of  settings. $\beta$ was a variable parameter and various experiments were conducted by setting its value with $\{0.95,0.85,0.75,0.65\}$.

\textbf{Quantitative results}. In this case, we show the classification results provided by the proposed deep embedded refined clustering (DERC) method depending on the value of $\beta$ (see Table \ref{tab:results_met_proposed}).

\begin{table}[ht]
\centering
\caption{Comparison of  the clustering effect of the proposed DECR based on different $\beta$ values.}
\label{tab:results_met_proposed}
\resizebox{8 cm}{!}{%
\begin{tabular}{lcccc}
\textbf{Method} & \textit{\textbf{ACC}} & \textit{\textbf{Error Rate(\%)}} & \textit{\textbf{FP}} & \textit{\textbf{FN}} \\ \hline
\multicolumn{1}{c}{DECR ($\beta=0.95$) } & 0.9708 & 2.92 & 4 & 0 \\ \hline
DECR ($\beta=0.85$)                             & 0.9781 & 2.19  & 3  & 0 \\ \hline
DECR ($\beta=0.75$)                          & \textbf{0.9927} & \textbf{0.73}  & \textbf{1}  & \textbf{0} \\ \hline
DECR ($\beta=0.65$)                          & 0.9854 & 1.4600  & 2  & 0 \\ \hline
\end{tabular}
}
\end{table}

\subsection{GSE50220 Series: Generalization ability of the DERC algorithm}

In this section, we expose the results for the prediction of an external test set, see Table \ref{tab:results_met2}.  The goal of this section is to demonstrate that the proposed DERC method could be valid to perform the feature extraction and classification from methylation data. Therefore, we made use of the GSE50220 series as an external test set to check the behaviour of the proposed methods with new breast cancer samples.

\begin{table}[ht]
\centering
\caption{Results obtained over the external dataset. Note that in this case, the input data corresponds to the GSE50220 series after selecting the CpG sites extracted with the statistical analysis in the GSE32393 series.}
\label{tab:results_met2}
\resizebox{8.2 cm}{!}{
\begin{tabular}{lcccc}
\textbf{Method} & \textit{\textbf{ACC}} & \textit{\textbf{Error Rate(\%)}} & \textit{\textbf{FP}} & \textit{\textbf{FN}} \\ \hline
\multicolumn{1}{c}{Input Data + K-means} & 0.6042 & 39.58 & 0 & 18 \\ \hline
AE + K-means                             & 0.8542 & 14.57   & 0  & 7 \\ \hline
DECR ($\beta=0.75$)                      & \textbf{0.9375} & \textbf{6.25}  & \textbf{0}  & \textbf{3} \\ \hline
\end{tabular}%
}
\end{table}

\section{Discussion} 
\label{5}

In this work, we present a deep embedded refined clustering approach to automatically detect patients suffering for cancer using DNA methylation data. In concrete, the proposed algorithm was evaluated using two breast methylation datasets.

As it can be observed in Table \ref{tab:results_met}, an optimal  data dimensionality reduction is essential  to improve the classification results when working with high-dimensional data. Using the K-means algorithm as non-supervised classifier, the dimensionality reduction carried out by the pre-trained AE on the input data (DNAm profiles obtained after statistical analysis) improves the ACC results from 0.6715 to 0.9343. However, with the VAE algorithm,  the ACC results do not improve, obtaining a value of 0.5693. As it can be seen in Fig. \ref{data_visualization}, the latent space of the VAE is centered around 0 due to the regularisation effect. This fact makes it impossible to distinguish between the different classes. Additionally, the reconstruction losses obtained by the VAE are higher than those reached by the AE (see Table \ref{tab:reconstruction_error}). Therefore, it can be concluded that the conventional autoencoder is the most suitable algorithm to reduce the DNAm dimensionality.

Moreover, regarding the comparison between classifying separately and jointly to the dimensionality reduction, ACC results show an improvement from 0.9343 to 0.9927 when the dimensionality reduction and the classification are optimized all at once. In table \ref{tab:results_met_proposed}, it can be observed the effect of the $\beta$ value on the classification results. In this way, it can be demonstrated that when the contribution of the reconstruction losses is too low (low $\beta$) or too high (high $\beta$), the ACC results are worse compared to a more balanced contribution. However, all the accuracy results shown in Table \ref{tab:results_met_proposed}, joint optimization, are higher than those obtained when applying K-means in the autoencoder latent space, separate optimization. Therefore, it is proven that when the classification is carried out at the same time as the dimensionality reduction is optimized, the best  results are obtained. 

This fact can also be demonstrated  when the results of the proposed method are compared to those obtained in the literature \cite{si2016learning,jazayeri2020breast}. As discussed in Section \ref{introduction}, in \cite{si2016learning}, the authors proposed a dimensionality reduction algorithm followed by several unsupervised classification algorithms. They applied their methods to the same breast cancer database used in this paper (GSE32393 series). Note that they obtained an error rate of 2.94  using a deep neural network (DNN) following a self-organizing feature map (SOM) compared to 0.73 obtained by the proposed method. Furthermore, their algorithm predicted 4 cancer examples as healthy, while our method only misclassifies one cancer sample. Therefore, it confirms that the proposed deep embedded refined clustering algorithm improves the results when it is applied to DNA methylation data. Additionally, the algorithm proposed in \cite{jazayeri2020breast} used the same breast cancer database  (GSE32393 series). In this case, they used a  Non-negative matrix factorisation (NMF) for dimensionality reduction following by supervised algorithms for classification. Their main limitation is that, according to the authors, the NMF algorithm cannot directly reduce the number of features (27,578) to a lower dimension than the number of samples (137). Therefore, they used a method called column-splitting in which they separated the original data into different matrices. They could not reduce the original data to a single latent space because they had to reduce each data matrix independently. In this way, the overall information of all original features is not taken into account. They used a K-fold for the algorithm validation and obtaining a 100 \% of accuracy when they used 900 and 2700 CpG sites. However, both resulting models were overfitted as it is demonstrated when they reduced the number of features to 540 and the accuracy dropped to 97.85 \%, lower than achieved with the proposed method. Additionally, the authors of \cite{jazayeri2020breast} claimed that it is important to reduce the number of features to a smaller space than the total number of examples. However, they were only able to reduce the features to 540 (due to NMF restrictions) which is about 5 times the number of examples they used to train their models. 

Furthermore, the results obtained by our proposed method on the external database (GSE50220 series) reported closely similar values to those reached in the primary set (GSE32393 series). This fact indicates that the proposed deep clustering model is perfectly applicable to other breast tissue databases (see Table \ref{tab:results_met2}). 

\section{Conclusion}
\label{6}
In this paper, a deep embedded refined clustering based on breast cancer classification using DNA methylation data has been presented. To the best of the authors' knowledge, no previous studies using DNA methylation are based on algorithms that can optimise the dimensionality reduction and the classification of the data at the same time. As demonstrated throughout the manuscript, the method proposed in this paper improves the results of algorithms using dimensionality reduction and subsequent classification.

The  proposed method allows the breast cancer classification using a latent space of only 10 features, which means a reduction in the dimensionality of 99.9637 \%.  The technology used in this study for data acquisition is the Illumina Infinium 27k Human DNA methylation Beadchip v1.2 which uses probes on the 27k array target regions of the human genome to measure methylation levels at 27,578 CpG dinucleotides in 14,495 genes. As verified through this work, many of the CpG sites obtained in the DNA methylation analysis are not relevant in the breast cancer classification. After ensuring model viability  with a larger breast cancer database, the CpG sites from which the level of methylation is obtained could be reduced decreasing the cost and time of methylation analysis. Therefore, this work could contribute to a faster and more effective diagnosis of breast cancer, improving cancer care and advancing the future of breast cancer research technologies.

From a technical perspective,  future lines of work will focus on adapting and applying the proposed method to identify and appropriately classify other challenging disorders, such as melanocytic tumours. In this way,  the general applicability of the model for the detection of different types of cancer could be demostrated.

\section*{Acknowledgments}
We  gratefully  acknowledge the support of NVIDIA Corporation with the donation of the Titan V GPU used for this research. NVIDIA Corporation had no role in study design, data collection and analysis, decision to publish or preparation of the manuscript.

\section*{Funding}

This work has received funding from Horizon 2020, the European Union's Framework Programme for Research and Innovation, under grant agreement No.  860627 (CLARIFY) and the Spanish Ministry of Economy and Competitiveness through project PID2019-105142RB-C21 (AI4SKIN).

\section *{Conflicts of interest}

The authors declare that they have no conflict of
interest.

\bibliographystyle{spmpsci}
\bibliography{mybibfile}

\begin{thebibliography}{10}
\providecommand{\url}[1]{{#1}}
\providecommand{\urlprefix}{URL }
\expandafter\ifx\csname urlstyle\endcsname\relax
  \providecommand{\doi}[1]{DOI~\discretionary{}{}{}#1}\else
  \providecommand{\doi}{DOI~\discretionary{}{}{}\begingroup
  \urlstyle{rm}\Url}\fi

\bibitem{akhavan2013dna}
Akhavan-Niaki, H., Samadani, A.A.: Dna methylation and cancer development:
  molecular mechanism.
\newblock Cell biochemistry and biophysics \textbf{67}(2), 501--513 (2013)

\bibitem{bibikova2011high}
Bibikova, M., Barnes, B., Tsan, C., Ho, V., Klotzle, B., Le, J.M., Delano, D.,
  Zhang, L., Schroth, G.P., Gunderson, K.L., et~al.: High density dna
  methylation array with single cpg site resolution.
\newblock Genomics \textbf{98}(4), 288--295 (2011)

\bibitem{du2010comparison}
Du, P., Zhang, X., Huang, C.C., Jafari, N., Kibbe, W.A., Hou, L., Lin, S.M.:
  Comparison of beta-value and m-value methods for quantifying methylation
  levels by microarray analysis.
\newblock BMC bioinformatics \textbf{11}(1), 587 (2010)

\bibitem{enguehard2019semi}
Enguehard, J., O’Halloran, P., Gholipour, A.: Semi-supervised learning with
  deep embedded clustering for image classification and segmentation.
\newblock IEEE Access \textbf{7}, 11093--11104 (2019)

\bibitem{esteller2008epigenetics}
Esteller, M.: Epigenetics in cancer.
\newblock New England Journal of Medicine \textbf{358}(11), 1148--1159 (2008)

\bibitem{variational_autoencoder}
Foster, D.: Generative deep learning: teaching machines to paint, write,
  compose, and play.
\newblock O'Reilly Media (2019)

\bibitem{GEO}
GEO: Epigenome analysis of breast tissue from women with and ithout breast
  cancer.
\newblock \url{http://www.ncbi.nlm.nih.gov/geo/query/acc.cgi?acc=gse32393}

\bibitem{guo2017deep}
Guo, X., Liu, X., Zhu, E., Yin, J.: Deep clustering with convolutional
  autoencoders.
\newblock In: International conference on neural information processing, pp.
  373--382. Springer (2017)

\bibitem{guo2018deep}
Guo, X., Zhu, E., Liu, X., Yin, J.: Deep embedded clustering with data
  augmentation.
\newblock In: Asian conference on machine learning, pp. 550--565 (2018)

\bibitem{hershey2016deep}
Hershey, J.R., Chen, Z., Le~Roux, J., Watanabe, S.: Deep clustering:
  Discriminative embeddings for segmentation and separation.
\newblock In: 2016 IEEE International Conference on Acoustics, Speech and
  Signal Processing (ICASSP), pp. 31--35. IEEE (2016)

\bibitem{jazayeri2020breast}
Jazayeri, N., Sajedi, H.: Breast cancer diagnosis based on genomic data and
  extreme learning machine.
\newblock SN Applied Sciences \textbf{2}(1), 3 (2020)

\bibitem{khwaja2018deep}
Khwaja, M., Kalofonou, M., Toumazou, C.: A deep autoencoder system for
  differentiation of cancer types based on dna methylation state.
\newblock arXiv preprint arXiv:1810.01243  (2018)

\bibitem{laird2010principles}
Laird, P.W.: Principles and challenges of genome-wide dna methylation analysis.
\newblock Nature Reviews Genetics \textbf{11}(3), 191--203 (2010)

\bibitem{liu2019dna}
Liu, B., Liu, Y., Pan, X., Li, M., Yang, S., Li, S.C.: Dna methylation markers
  for pan-cancer prediction by deep learning.
\newblock Genes \textbf{10}(10), 778 (2019)

\bibitem{maaten2008visualizing}
Maaten, L.v.d., Hinton, G.: Visualizing data using t-sne.
\newblock Journal of machine learning research \textbf{9}(Nov), 2579--2605
  (2008)

\bibitem{martorell2019deep}
Martorell-Marug{\'a}n, J., Tabik, S., Benhammou, Y., del Val, C., Zwir, I.,
  Herrera, F., Carmona-S{\'a}ez, P.: Deep learning in omics data analysis and
  precision medicine.
\newblock In: Computational Biology [Internet]. Codon Publications (2019)

\bibitem{min2018survey}
Min, E., Guo, X., Liu, Q., Zhang, G., Cui, J., Long, J.: A survey of clustering
  with deep learning: From the perspective of network architecture.
\newblock IEEE Access \textbf{6}, 39501--39514 (2018)

\bibitem{prasetio2019deep}
Prasetio, B.H., Tamura, H., Tanno, K.: A deep time-delay embedded algorithm for
  unsupervised stress speech clustering.
\newblock In: 2019 IEEE International Conference on Systems, Man and
  Cybernetics (SMC), pp. 1193--1198. IEEE (2019)

\bibitem{sharma2010epigenetics}
Sharma, S., Kelly, T.K., Jones, P.A.: Epigenetics in cancer.
\newblock Carcinogenesis \textbf{31}(1), 27--36 (2010)

\bibitem{si2016learning}
Si, Z., Yu, H., Ma, Z.: Learning deep features for dna methylation data
  analysis.
\newblock IEEE Access \textbf{4}, 2732--2737 (2016)

\bibitem{tian2019clustering}
Tian, T., Wan, J., Song, Q., Wei, Z.: Clustering single-cell rna-seq data with
  a model-based deep learning approach.
\newblock Nature Machine Intelligence \textbf{1}(4), 191--198 (2019)

\bibitem{titus2018unsupervised}
Titus, A.J., Wilkins, O.M., Bobak, C.A., Christensen, B.C.: Unsupervised deep
  learning with variational autoencoders applied to breast tumor genome-wide
  dna methylation data with biologic feature extraction.
\newblock bioRxiv p. 433763 (2018)

\bibitem{tsou2002dna}
Tsou, J.A., Hagen, J.A., Carpenter, C.L., Laird-Offringa, I.A.: Dna methylation
  analysis: a powerful new tool for lung cancer diagnosis.
\newblock Oncogene \textbf{21}(35), 5450--5461 (2002)

\bibitem{xie2016unsupervised}
Xie, J., Girshick, R., Farhadi, A.: Unsupervised deep embedding for clustering
  analysis.
\newblock In: International conference on machine learning, pp. 478--487 (2016)

\bibitem{yuvaraj2013efficient}
Yuvaraj, N., Vivekanandan, P.: An efficient svm based tumor classification with
  symmetry non-negative matrix factorization using gene expression data.
\newblock In: 2013 International Conference on Information Communication and
  Embedded Systems (Icices), pp. 761--768. IEEE (2013)

\bibitem{zhang2015predicting}
Zhang, W., Spector, T.D., Deloukas, P., Bell, J.T., Engelhardt, B.E.:
  Predicting genome-wide dna methylation using methylation marks, genomic
  position, and dna regulatory elements.
\newblock Genome biology \textbf{16}(1), 14 (2015)

\end{thebibliography}

\end{document}